\documentclass[conference]{IEEEtran}
\IEEEoverridecommandlockouts
\usepackage{cite}
\usepackage{amsmath,amssymb,amsfonts}
\usepackage{algorithmic}
\usepackage{graphicx}
\usepackage{textcomp}
\usepackage{xcolor}
\usepackage{balance}
\usepackage{url} 
\begin{document}
\title{Channel-Informed Neural Network for Physical Layer Key Generation}

% author names and affiliations
% use a multiple column layout for up to three different
% affiliations

%\makeatletter
%\newcommand{\linebreakand}{%
%  \end{@IEEEauthorhalign}
%  \hfill\mbox{}\par
  %\mbox{}\hfill\begin{@IEEEauthorh%align}
%}
%\makeatother

\author{
Jose Angel Sanchez Viloria$^*$, George Sklivanitis$^*$, Dimitris Pados$^*$, and Elizabeth Serena Bentley$^{\dagger}$%
\thanks{Distribution A. Approved for public release: Distribution Unlimited: AFRL-2026-3584 on 10 Aug 2026. 
% Reviewer comment: place “Distribution A. Approved for public release: Distribution Unlimited: AFRL-2026-3584 on 10 Aug 2026" somewhere on the front page
This work was supported in part by NSF Grants EEC-2133516, CNS-2117822, ITE-2226392 and by the Air Force Research Laboratory Grants FA8750-21-F-1012, FA8750-20-C-1021.}
\\
$^*$Center for Connected Autonomy and AI, Florida Atlantic University, Boca Raton, FL, USA\\
\{\texttt{josesanchez2019, gskivanitis, dpados}\}@fau.edu\\
$^{\dagger}$Air Force Research Laboratory, Rome, NY, USA\\
\{\texttt{elizabeth.bentley.3}\}@us.af.mil\\

}

% make the title area
\maketitle

% \renewcommand{\thefootnote}{}
% \footnotetext{
% This work was supported in part by NSF Grants EEC-2133516 and CNS-2117822, and by the Air Force Research Laboratory Grant FA8750-21-F-1012.
% }

\begin{abstract}
Physical-layer key generation (PKG) enables wireless devices to establish shared %cryptographic 
keys from reciprocal channel observations without directly exchanging the key. This capability is attractive for edge networks, where distributed and resource-constrained devices may require lightweight key establishment with limited access to centralized infrastructure.
We introduce a channel-informed neural network for PKG that derives binary key features directly from received IQ measurements while explicitly grounding the learned representation in the underlying multipath channel.
The proposed multi-task recurrent neural network jointly learns reciprocity-preserving binary features and an auxiliary channel estimate using a training objective that combines deep metric learning with channel-informed supervision.
Structured channel sounding enables channel estimation from over-the-air measurements, while Sionna-RT ray tracing is used to augment training with additional propagation conditions.
We evaluate the framework using indoor and outdoor software-defined-radio measurements collected on the POWDER radio testbed. Across all evaluated scenarios, the proposed model produces lower bit disagreement for reciprocal Alice–Bob observations than for Eve-related observations. Ray-traced data augmentation substantially improves key diversity, increasing the unique-key rate to $0.94$, $0.99$, and $0.99$ across the indoor and two outdoor scenarios, respectively. Successfully reconciled channel-informed keys pass the selected NIST randomness tests prior to SHA-3 privacy amplification. The results demonstrate the potential of channel-informed representation learning for decentralized wireless key establishment while highlighting an important tradeoff between key diversity and reconciliation reliability.
\end{abstract}


% Start with problem, why it is important, description of literature, and our method

\section{Introduction}

Secure key establishment is a foundational requirement for edge communications, where autonomous platforms, distributed sensors, and resource-constrained wireless devices may operate over contested or untrusted networks with intermittent access to centralized infrastructure. 
Although common mechanisms for encryption depend on the secure establishment and timely renewal of shared secret keys, key distribution infrastructure, or trusted key-management entities, which may introduce computational, communication, and operational overhead in denied, disconnected, intermittent, or limited-bandwidth environments. 
%Reviewer Comment: Scrub and Remove cryptography and associated terms (symmetric/asymmetric/pki/ in multiple locations (This is on CCL), Remove tactical and deployment (multiple occurrences), Pg 1 remove “attacks”, Remove reference 2 on quantum cryptography – (On CCL) – I tried to find one that was related rather than change all the reference numbers by removing it., Remove any mention of SHA (Page 4 – “by hashing the reconciled bit sequence with the SHA-3 hash function key = SHA3 (yi) = SHA3 (byj) , key ∈ {0, 1}512.  The resulting 512-bit digest is used as the final key.”   Which is edited to “hashing the reconciled binary feature vector with privacy amplification to produce a key.”  Interestingly this variable Key is not used after this on page 4 so she read that as not having any technical impact and could be replaced generically.  
% Although symmetric cryptography provides an efficient mechanism for protecting mission-critical data, its effectiveness depends on the secure establishment and timely renewal of shared secret keys. Conventional key-establishment mechanisms commonly rely on asymmetric cryptography, public-key infrastructure...^
These constraints motivate complementary approaches that enable decentralized and lightweight key establishment directly between wireless edge devices while reducing dependence on persistent connectivity to centralized services \cite{8715341}.
%Symmetric cryptography remains a widely used mechanism for confidentiality, but the corresponding key establishment process often relies on asymmetric cryptographic protocols, public-key infrastructure, or trusted key-management entities, which may introduce computational and operational overhead \cite{8715341}. In addition, the emergence of quantum computing motivates the investigation of complementary and lightweight mechanisms for key establishment in wireless systems \cite{ChenEtAl2016}.

Physical-layer key generation (PKG) provides such a complementary mechanism by exploiting the reciprocal, time-varying, and spatially decorrelated characteristics of wireless propagation channels \cite{7393435}. Two trusted network users, conventionally denoted as Alice and Bob, exchange channel probes within the channel coherence time and independently derive correlated channel observations. These observations are transformed into preliminary binary sequences, residual disagreements are corrected through information reconciliation, and privacy amplification is applied to generate a shared cryptographic key \cite{11106373,1193795,1193794,1193793,9598159,10213440}. Because the key is derived from locally observed channel characteristics rather than directly exchanged over the network, PKG is particularly attractive for edge devices that require lightweight and rapidly renewable key material.

%Physical-layer key generation (PKG) offers a complementary approach by exploiting the reciprocity, temporal variability, and spatial decorrelation of wireless channels to allow two legitimate nodes, Alice and Bob, to derive shared secret keys without directly exchanging the key material \cite{7393435}.
%In a typical PKG protocol, Alice and Bob exchange channel probes within the coherence time of the channel, extract correlated channel observations, quantize these observations into binary sequences, reconcile residual bit disagreements, and apply privacy amplification to obtain a cryptographic key \cite{11106373,1193795,1193794,1193793,9598159,10213440}. 
%The effectiveness of this process depends strongly on the ability of the feature extraction stage to preserve the common randomness observed by Alice and Bob while reducing the correlation between the legitimate features and those available to a passive eavesdropper.
\begin{figure*}[t]
\centering
\includegraphics[width=\textwidth]{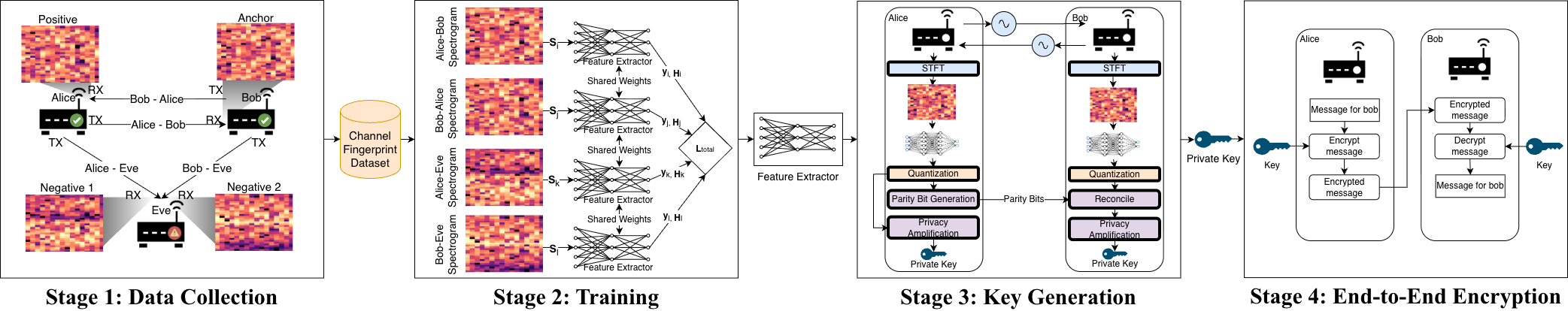}
\caption{Data collection, model training, and testing of proposed PKG protocol for dynamic data encryption.}
\label{fig:SystemModel}
\end{figure*}

Prior PKG methods commonly derive shared randomness directly from channel-state information (CSI), received-signal strength (RSS), channel-frequency response, or related propagation measurements \cite{KZeng,HLiu,SMathur}. Recent work has increasingly focused on improving the reliability and deployability of PKG under non-ideal operating conditions including adaptive probing \cite{YWei} interpolation and decorrelation transformations \cite{NPatwari}, mismatch-aware key generation \cite{STAli}. 
%, and neural-network-assisted reciprocity learning or bit correction \cite{FDD_OFDM_MultiEnv,  10.1145/3522783.3529526,NN_PHY_Key_Exchange}. 
Deep-learning-based methods have been proposed to construct reciprocal features in frequency-division-duplex (FDD) systems and to adapt learned mappings across previously unseen propagation environments \cite{FDD_OFDM_MultiEnv,  10.1145/3522783.3529526,NN_PHY_Key_Exchange}. 
Experimental studies on resource-constrained radios have also shown that practical hardware limitations and time-varying channel conditions can significantly degrade CSI reciprocity, motivating signal-processing-based reconstruction techniques \cite{10.1016/j.cose.2025.104423}. Other recent studies address FDD key generation and hardware-induced impairments through pairwise or loop-back probing mechanisms \cite{10693595, 10.1109/TIFS.2025.3564064}, while emerging research has examined adversarial-learning defenses against malicious reconfigurable-intelligent-surface %attacks (Pg 1 remove “attacks”)
\cite{10856736}, neural-enhanced reconciliation \cite{11153007}, and large-language-model-assisted probing for vehicular networks \cite{LLMKey}.
%
%Prior PKG methods commonly rely on channel state information (CSI), received signal strength (RSS), or related channel measurements as direct sources of randomness \cite{KZeng,HLiu,SMathur}. 
%Other works improve key reliability through adaptive probing \cite{YWei}, interpolation and decorrelation transformations \cite{NPatwari}, mismatch-aware key generation \cite{STAli}, and neural-network-assisted reciprocity learning or bit correction \cite{FDD_OFDM_MultiEnv,10.1145/3522783.3529526,NN_PHY_Key_Exchange}. 
%Recent work has also explored learning-based and large-language-model-assisted mechanisms for wireless key generation \cite{LLMKey}. 
%
%While these approaches demonstrate the potential of data-driven PKG, many learning-based methods treat the neural feature extractor primarily as an end-to-end mapping from channel observations to embeddings or binary sequences. As a result, the learned representation may not be explicitly constrained to preserve physically meaningful channel structure during training.
%
Collectively, these efforts demonstrate the value of data-driven methods for improving key agreement under practical channel conditions. However, much of the existing literature either operates on precomputed channel descriptors or optimizes an individual stage of the PKG pipeline. Comparatively less attention has been given to neural feature extractors that operate on raw IQ observations while being explicitly constrained to preserve the underlying multipath-channel structure during training.

In this paper, we introduce a channel-informed learning framework for PKG that strengthens the connection between learned security features and the underlying wireless propagation environment.
 Building on our prior AI-assisted PKG work \cite{11443725}, the proposed approach moves beyond purely data-driven feature extraction by combining structured channel probing with auxiliary channel-estimation supervision and ray-traced propagation data. Specifically,  our proposed physical layer key generation design encourages the neural network to preserve channel-dependent structure while learning binary features that remain more consistent for reciprocal Alice–Bob observations than for Eve-related observations. The resulting framework supports decentralized key establishment from raw IQ measurements and is evaluated as an end-to-end pipeline comprising feature extraction, information reconciliation, privacy amplification, and dynamic data encryption.

The primary contributions of this work are as follows:
\begin{itemize}
\item %Channel-informed key-feature learning from raw IQ measurements: 
We introduce a multi-task neural network framework that jointly learns reciprocity-preserving binary features and an auxiliary representation of the multipath channel. Unlike approaches that operate only on precomputed channel descriptors or optimize embedding similarity alone, the proposed method explicitly grounds learned key features in measurable propagation structure.
\item %Digital-twin-augmented PKG training: 
We integrate over-the-air channel measurements with ray-traced wireless scenarios to broaden the propagation conditions represented during training. Across the evaluated indoor and outdoor test settings, this augmentation increases the unique-key rate, substantially improving key diversity relative to the baseline \cite{11443725}.
\item %Experimental characterization of the agreement–diversity tradeoff: 
We evaluate the proposed framework using synthetic data, indoor and outdoor software-defined-radio (SDR) measurements. The results show that channel-informed training consistently separates reciprocal legitimate-channel observations from eavesdropper observations, while also exposing a practical tradeoff between increased key diversity and reconciliation success. This tradeoff identifies an important design dimension for robust PKG systems operating at the edge.
\end{itemize}

\section{System Model}

We aim to independently generate a shared secret key from reciprocal wireless-channel observations between two trusted network nodes, Alice and Bob. % Reviewer Comment: The manuscript should explicitly state that the public reconciliation channel must be authenticated. Physical-layer key generation alone does not prevent an active adversary from modifying reconciliation messages or impersonating one of the legitimate devices. (We state this here) 
Alice and Bob exchange known channel probes in a time-division duplex (TDD) fashion, within the channel coherence time, so their forward and reverse channel observations are expected to be highly correlated but not identical due to noise, hardware impairments, and synchronization offsets \cite{7393435,8715341}. A passive eavesdropper, Eve $(E)$, is informed about the key generation protocol and monitors the probes exchanged between Alice and Bob, but does not actively modify the channel.  %The considered links are $(u,v)\in\{(A,B),(B,A),(A,E),(B,E)\}$.

We consider channel probing with a BPSK-modulated code sequence based on the Galois Linear Feedback Shift Register (GLFSR). %We consider a total of four links between Alice-Bob, Bob-Alice, Alice-Eve, and Bob-Eve. % denoted as $(u,v)\in\{(A,B),(B,A),(A,E),(B,E)\}$. %Each probe is generated from a GLFSR binary sequence
%\begin{flalign*}
%    b[n]\in\{0,1\}, \qquad n=0,\ldots,N_p-1,
%\end{flalign*}
%where $N_p$ is the probe length. The bits are mapped to BPSK symbols using the antipodal mapping
\begin{flalign*}
   s[n] = 1-2b[n] \in \{+1,-1\},
   \qquad n=0,\ldots,N_p-1,
\end{flalign*}
%where $s[n]\in\mathbb{C}$ denotes the transmitted complex baseband probe sample with zero imaginary component \cite{ProakisDigitalComms}.
%
The received baseband signal at Alice or Bob or Eve after carrier frequency downconversion and sampling at integer multiples
of the sampling period $T_s$ is written as follows
\begin{align}
\nonumber
x[n] = \sum_{p=1}^{P} h_p s(nT_s - \tau_p)+w[n], \quad n = 0, \dots, N_p + P- 1
\end{align}
% Reviewer Comment: Something is strange about the indexing of n throughout the paper. On page two, first column, n ranges from 0 to N_p + P -1, but in other places it only goes up to N_p -1. The discussion on page 5 seems to indicate the added P is intentional, but why it should be there is not clear. Similarly, Section III. A. paragraph 1 indicates a range of 0 to N_p + P - 2. It's not clear why this should be one less, (regardless of whether P is present). Indeed, the equation below seems to quibble between N_P + P - 1 (an explicit range stated for n) and N_P - 1 (the bounds on the summation). If this is all intentional, a better description of why would help the reader. (Described here for the difference between N_p and P)
where $p = 1, \dots, P$ is the number of channel paths, $h_p$ and $\tau_p$ denote the attenuation and delay of the $p$-th path, respectively
%Each wireless link is modeled as an $L$-tap discrete-time complex baseband multipath channel. The received channel probe at each node is given by %$v$ from transmitter $u$ is
%\begin{flalign*}
%    x[n]
%    =
%    \sum_{\ell=0}^{L-1}
%    h_{u,v}[\ell]s[n-\ell]
%    +
%    w_{u,v}[n],
%\end{flalign*}
%where $x_{u,v}[n]\in\mathbb{C}$, $n=0,\ldots,N_r-1$, is the $n$-th received complex baseband sample, 
$s[n]$, $n=0, 1, \dots, N_p-1$, is the $n$-th symbol of the transmitted GLFSR code sequence,  %$h[\ell]\in\mathbb{C}$ is the $\ell$-th channel tap, $L$ is the number of resolvable taps, 
$w[n]\sim\mathcal{CN}(0,\sigma^2)$ is complex additive Gaussian noise with variance $\sigma^2$. %and $N_r=N_p+L-1$ is the received sequence length. %for linear convolution \cite{ProakisDigitalComms,GoldsmithWireless}.
% In vector form, \eqref{eq:rx_signal_model} is written as
% \begin{flalign*}
%     \mathbf{x}_{u,v}
%     =
%     \mathbf{A}_{s}\mathbf{h}_{u,v}
%     +
%     \mathbf{w}_{u,v},
% \end{flalign*}
% where $\mathbf{x}_{u,v}\in\mathbb{C}^{N_r}$ is the received vector, $\mathbf{h}_{u,v}\in\mathbb{C}^{L}$ is the channel-tap vector, $\mathbf{w}_{u,v}\in\mathbb{C}^{N_r}$ is the noise vector, and $\mathbf{A}_{s}\in\mathbb{C}^{N_r\times L}$ is the Toeplitz convolution matrix formed from the known probe sequence $s[n]$.

Since the channel probe sequence $s[n], n=0, 1, \dots, N_p-1$ is assumed to be known at the receiver, the least squares (LS) channel estimate is given by
\begin{flalign*}
    \widehat{\mathbf{h}}^{\mathrm{LS}}
    =
    \arg\min_{\mathbf{h}}
    \left\|
    \mathbf{x}
    -
    \mathbf{S}\mathbf{h}
    \right\|_2^2
\end{flalign*}
where $\mathbf{x}\in\mathbb{C}^{N_p+P-1}$ is the received probe signal samples, $\mathbf{h}\in\mathbb{C}^{P}$ is the channel impulse response vector, and $\mathbf{S}\in\mathbb{C}^{N_p + P-1\times P}$ is the Toeplitz convolution matrix formed from the known channel probing sequence.
% When $\mathbf{A}_{s}^{H}\mathbf{A}_{s}$ is nonsingular, the LS solution is
% \begin{flalign*}
%     \widehat{\mathbf{h}}_{u,v}^{\mathrm{LS}}
%     =
%     \left(
%     \mathbf{A}_{s}^{H}\mathbf{A}_{s}
%     \right)^{-1}
%     \mathbf{A}_{s}^{H}
%     \mathbf{x}_{u,v},
% \end{flalign*}
% where $(\cdot)^H$ denotes Hermitian transpose \cite{KayEstimation}. 
%The estimates $\widehat{\mathbf{h}}_{u,v}^{\mathrm{LS}}$ are used as channel-informed training targets for the feature extractor.
%
%For legitimate probing, 

%We also generate synthetic channel data using Sionna-RT. 
We also consider a ray tracer for simulating radio wave propagation. Specifically, we rely on Sionna-RT \cite{SionnaRTDocs} to compute channel impulse responses for ray-traced propagation paths. %with complex path coefficients and delays. 
Given the continuous-delay channel impulse response below
\begin{flalign*}
    h^{\mathrm{RT}}(\tau)
    =
    \sum_{p=1}^{P}
    a_p\delta(\tau-\tau_p)
\end{flalign*}
where $a_p$, $\tau_p$ are the complex channel coefficient and delay of the $p$-th channel path, respectively, $\tau$ is the continuous delay, and $\delta(\cdot)$ is the Dirac delta function \cite{SionnaRTDocs,GoldsmithWireless}, the $l$-th channel tap at sample instance $n$ is computed as follows 
%we  obtain discrete taps, the ray-traced response is projected onto the sampling grid as
\begin{flalign*}
     h^{\mathrm{RT}}_{n,\ell}
     =
     \sum_{p=1}^{P}
     a_p\left(\frac{n}{B}\right)
     \operatorname{sinc}(\ell - B \tau_p),
     %(\ell - B \tau_p),
     \qquad
     \ell=l_{\rm{min}},\ldots,l_{\rm{max}}
 \end{flalign*}
 where $l_{\rm{min}}$, $l_{\rm{max}}$ are the smallest and largest time-lag for the discrete complex baseband-equivalent channel, respectively and $B$ is the bandwidth to which the channel impulse response is limited \cite{ProakisDigitalComms,SionnaRTDocs}. % $T_s$ is the sampling period and $g(\cdot)$ is the effective pulse-shaping and receiver-filter response  

We expect that both ray-tracer computed and over-the-air estimated channel impulse responses between Alice--Bob and Bob--Alice are approximately reciprocal while Eve observes different channels that are expected to be less correlated with the  Alice--Bob/Bob--Alice channel when Eve is spatially separated from Alice and Bob \cite{7393435}. In Section III, we describe how the channel impulse response information (acquired either by ray-tracing or channel estimation) is used for grounding the training of neural-network based binary feature extractors who operate directly on received IQ spectrograms.  %neural-network based feature extraction 
%The forward and reverse channels are approximately reciprocal,
%$\mathbf{h}_{A,B}\approx\mathbf{h}_{B,A}$ and $\mathbf{h}_{A,B}\neq\mathbf{h}_{B,A}$ while Eve observes different channels $\mathbf{h}_{A,E}$ and $\mathbf{h}_{B,E}$ that are expected to be less correlated with the  Alice--Bob channel when Eve is spatially separated from Alice and Bob \cite{7393435}. LS channel estimates are used as training targets for the proposed channel-informed neural network feature extractor which will be described in Section III.
%The Sionna-RT generated channel taps $\mathbf{h}_{u,v}^{\mathrm{RT}}$ are used as additional channel-informed training targets to augment the number of propagation conditions used during training of the proposed channel-informed neural network feature extractor.

\section{Channel-Informed Learning PKG Framework}
In this section, we describe in detail the architecture of our proposed channel-informed learning PKG framework. Our goal is to generate reciprocity-preserving binary keys directly from IQ spectrograms that are grounded to the physics of the wireless channel between a pair of trusted network nodes. Figure \ref{fig:SystemModel} depicts the end-to-end workflow for data collection, model training and testing of the proposed framework.

\subsection{Signal Pre-processing}

For each received channel probe, the complex baseband IQ samples are first normalized before converted into a time--frequency representation. Let 
$
    \mathbf{x} =
    \left[
    x[0], x[1], \ldots, x[N_p+P-1]
    \right]^T
    \in \mathbb{C}^{N_p+P-1}
$
% \begin{flalign*}
%     \mathbf{x} =
%     \left[
%     x[0], x[1], \ldots, x[L_s-1]
%     \right]^T
%     \in \mathbb{C}^{L_s}
% \end{flalign*}
denote the received IQ sample vector associated with one channel probe, where $N_p$ is the number of samples used for feature extraction (which in this special case is equal to number of symbols of the GLFSR code sequence as we consider $1$ sample per symbol). 
%Reviewer comment: sample-per-symbol -> sample per symbol
To reduce the effect of received-power variations across links and probing rounds, we apply root-mean-square (RMS) normalization:
\begin{flalign*}
    \widetilde{x}[n]
    =
    \frac{x[n]}
    {
    \sqrt{
    \frac{1}{N_p}
    \sum_{n=0}^{N_p-1}
    |x[n]|^2
    }
    +
    \epsilon
    },
    \quad
    n=0,\ldots,N_p+P-1
\end{flalign*}
where $\epsilon>0$ is a small constant used for numerical stability.

%The normalized IQ sample sequence is then transformed using the short-time Fourier transform (STFT). 
The normalized IQ sample sequence is divided into $M$ overlapping segments using a Hanning window of length $N_F$ and hop size $R$ samples, resulting in $M = 1+((N_p+P-N_F)/R)$ segments.
% \begin{flalign*}
%     M =
%     1+
%     \left\lfloor
%     \frac{L_s-N_F}{R}
%     \right\rfloor .
% \end{flalign*}
The $m$-th segment of the signal is given by
\begin{flalign*}
    \widetilde{x}_{m}[n]
    =
    \widetilde{x}[n+mR],
    \qquad
    n=0,\ldots,N_F-1,
\end{flalign*}
for $m=0,\ldots,M-1$. We define a Hanning window of $N_F$ samples as follows
$w[n]= 0.5(1-\cos(2\pi n/(N_F-1))), n = 0,1,...,N_F-1.$
% \begin{flalign*}
%     w[n]
%     =
%     0.5
%     \left(
%     1-\cos\left(\frac{2\pi n}{N_F-1}\right)
%     \right),
%     \qquad
%     n=0,\ldots,N_F-1,
% \end{flalign*}
and apply it to each of the $M$ segments of the signal to reduce spectral leakage. The discrete Fourier Transform is then applied to each windowed segment as follows
\begin{flalign*}
    X_{m,k}
    =
    \sum_{n=0}^{N_F-1}
    \widetilde{x}_{m}[n]w[n]
    e^{-j2\pi nk/N_F},
    \quad
    k=0,\ldots,N_F-1,
\end{flalign*}
where $X_{m,k}$ is the $(m,k)$-th element of the short-time Fourier Transform (STFT) complex matrix $\mathbf{X}\in\mathbb{C}^{M\times N_F}$.
The resulting STFT matrix $\mathbf{X}\in\mathbb{C}^{M\times N_F}$ represents the time--frequency structure of the received probe. Since the proposed neural-network based feature extractor operates on real-valued inputs, the complex STFT coefficients are separated into their real and imaginary components
$
    \mathbf{S}
    =
    \left[
    \Re\{\mathbf{X}\},
    \Im\{\mathbf{X}\}
    \right]
    \in
    \mathbb{R}^{M\times N_F\times 2}
$.
%The tensor $\mathbf{S}$ is used as the input to the channel-informed feature extractor.

\subsection{Neural Network Architecture}

\begin{figure}[t]
\centering
\includegraphics[width=0.8\columnwidth]{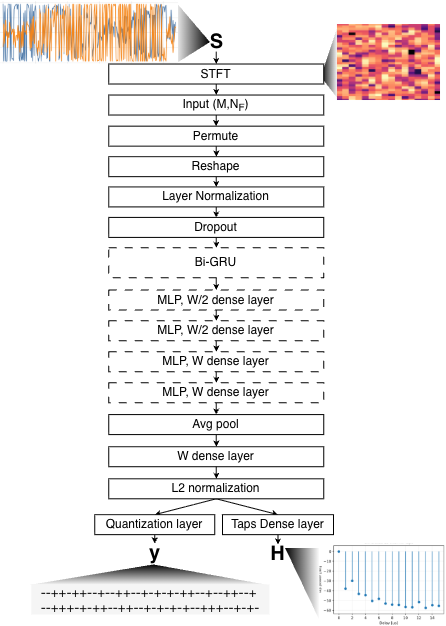}
\caption{Channel-informed GRU feature extractor with binary-feature and channel-estimation heads.}
\label{fig:NN}
\end{figure}

Figure \ref{fig:NN} depicts the neural network architecture for the proposed channel-informed binary feature extractor. We use a multi-task recurrent neural network (RNN) to jointly learn reciprocity-preserving key features and channel-estimation structure. %The input IQ probe is first unit-RMS normalized and converted performing STFT to compute the spectrogram $\mathbf{S}\in\mathbb{R}^{M\times N_F\times2}$
% \begin{flalign*}
%     \mathbf{S}\in\mathbb{R}^{M\times N\times2},
% \end{flalign*}
%where the frequency and time dimensions are given by $N_F=32$ and $M=16$, respectively. 
The input spectrogram ${\bf S}$ is interpreted as a sequence of $M$ time-frequency slices, where each time step contains $N_F$ features. In this work we consider $N_F=32$ and $M=16$. After reshaping and normalization, the sequence is processed by a bidirectional gated recurrent unit (GRU) layer with $384$ hidden units per direction \cite{GRU,650093}. This allows the encoder to aggregate temporal context from both forward and backward directions across the received probe response.

The GRU output is refined using four dense normalization blocks and then aggregated into a fixed-length latent representation. Specifically, the dense blocks project the sequence features to a compact embedding, followed by global average pooling and a dense projection to obtain $\mathbf{z}\in\mathbb{R}^{W}$, where $W$ is the output embedding size and denotes the preferred length of the encryption key.
% \begin{flalign*}
%     \mathbf{z}\in\mathbb{R}^{128}.
% \end{flalign*}
% Unlike the previous RNN feature extractor, the final latent vector is not passed through a sigmoid activation. 
% Instead, we apply L2 normalization,
% \begin{flalign*}
%     \widetilde{\mathbf{z}}
%     =
%     \frac{\mathbf{z}}{\|\mathbf{z}\|_2+\epsilon},
% \end{flalign*}
% where $\epsilon>0$ is used for numerical stability. 
The
final embedding is passed through an L2 normalization layer. The normalized latent vector $\widetilde{\mathbf{z}}$ is shared by two task-specific heads.
The first head produces the binary feature vector used for key generation. It applies mean-centered sign quantization %with a straight-through estimator (STE):
\begin{flalign*}
    \mathbf{y}
    =
    \mathrm{sign}
    \left(
    \widetilde{\mathbf{z}}
    -
    \frac{1}{W}\sum_{w=1}^{W}\widetilde{z}_w
    \right),
    \qquad
    \mathbf{y}\in\{-1,+1\}^{W}
\end{flalign*}
where $W=128$ is the size of the key we selected for this work. During backpropagation, we apply straight-through estimator (STE) which approximates the gradient of the non-differentiable sign operation, enabling end-to-end training of the binary feature head.
The second head performs auxiliary channel estimation from the same continuous normalized embedding before quantization. It applies a linear dense layer of size $2P$ followed by a reshape operation $\widehat{\mathbf{H}} \in \mathbb{R}^{P\times2}$,
% \begin{flalign*}
%     \widehat{\mathbf{H}}
%     \in
%     \mathbb{R}^{L_h\times2},
% \end{flalign*}
where $P=16$ in the current implementation (i.e., $16$ propagation paths). The final dimension stores the real and imaginary components of the estimated complex channel taps. During training, the binary feature head is optimized with the metric-learning objective, while the channel-estimation head is optimized with the channel loss (as described in more detail in Section III-C). During key generation, only the binary feature head is required.

\subsection{Channel-Informed Deep Metric Learning}

The proposed feature extractor is trained to satisfy two complementary objectives. First, it  learns an embedding space suitable for PKG, where reciprocal channel observations from Alice--Bob and Bob--Alice are mapped close to each other, while observations involving Eve are mapped farther away. Second, the learned representation should remain consistent with the underlying multipath channel response computed from either the known GLFSR-based channel probe or from the ray-tracer. This second objective introduces a channel-informed constraint that plays a role similar to physics-informed regularization i.e., the network is not trained only from embedding distances, but is also guided by channel impulse response targets derived from either the probing signal model or the ray-tracer.

The data loss is based on a quadruplet metric-learning objective \cite{QuadrupletLoss}. Let a training quadruplet be denoted by 
% $\mathcal{Q}_p = \left(\mathbf{S}_{i,p},\mathbf{S}_{j,p},\mathbf{S}_{k,p},\mathbf{S}_{l,p}\right)$
\begin{flalign*}
    \mathcal{Q}_c =
    \left(
    \mathbf{S}_{i,c},
    \mathbf{S}_{j,c},
    \mathbf{S}_{k,c},
    \mathbf{S}_{l,c}
    \right),
\end{flalign*}
where $\mathbf{S}_{i,c}$, $\mathbf{S}_{j,c}$, $\mathbf{S}_{k,c}$, and $\mathbf{S}_{l,c}$ correspond to the Alice--Bob, Bob--Alice, Alice--Eve, and Bob--Eve spectrograms for the $c$-th bidirectional channel probing round, respectively. The shared neural network maps each spectrogram to a continuous feature vector
$\mathbf{y}_{r,c}=f_{\theta}(\mathbf{S}_{r,c}) \in \mathbb{R}^{W}$, $ r\in\{i,j,k,l\}$
% \begin{flalign*}
%     \mathbf{y}_{r,p}=f_{\theta}(\mathbf{S}_{r,p}) \in \mathbb{R}^{D},
%     \qquad
%     r\in\{i,j,k,l\},
% \end{flalign*}
where $W$ is the embedding dimension. A second channel-estimation head maps the same learned representation to an estimated complex channel-tap vector $\widehat{\mathbf{h}}_{r,c}=g_{\theta}(\mathbf{S}_{r,c}) \in \mathbb{C}^{P}$,
% \begin{flalign*}
%     \widehat{\mathbf{h}}_{r,p}=g_{\theta}(\mathbf{S}_{r,p}) \in \mathbb{C}^{L},
% \end{flalign*}
where $P$ is the number of channel taps. The corresponding reference channel vector $\mathbf{h}_{r,c}$ is obtained from either the LS channel estimate acquired by the GLFSR-based channel probe or from the Sionna-RT ray-tracer.

As a result the proposed training objective combines a data-driven metric-learning loss with a channel-informed loss
\begin{flalign*}
    \mathcal{L}_{\mathrm{Total}}
    =
    \mathcal{L}_{\mathrm{Data}}
    +
    \lambda_{\mathrm{ch}}
    \mathcal{L}_{\mathrm{Channel}}
\end{flalign*}
where $\lambda_{\mathrm{ch}}\geq 0$ controls the contribution of the channel constraint.

 More specifically, for the metric-learning loss we use the squared Euclidean distance between binary feature vectors,
$d(\mathbf{y}_{a},\mathbf{y}_{b})= {\|\mathbf{y}_{a}-\mathbf{y}_{b}\|_2^2} /{W},$
% \begin{flalign*}
%     d(\mathbf{y}_{a},\mathbf{y}_{b})
%     =
%     \frac{\|\mathbf{y}_{a}-\mathbf{y}_{b}\|_2^2}{D},
% \end{flalign*}
and define
\begin{flalign*}
\mathcal{L}_{\mathrm{Data}}
=
\frac{1}{C}
\sum_{c=1}^{C}
\Big[
&
\max\left(
d(\mathbf{y}_{i,c},\mathbf{y}_{j,c})
-
d(\mathbf{y}_{i,c},\mathbf{y}_{k,c})
+
\alpha,
0
\right)
\nonumber\\
+
&
\max\left(
d(\mathbf{y}_{i,c},\mathbf{y}_{j,c})
-
d(\mathbf{y}_{j,c},\mathbf{y}_{l,c})
+
\beta,
0
\right)
\nonumber\\
+
&
\max\left(
d(\mathbf{y}_{i,c},\mathbf{y}_{j,c})
-
d(\mathbf{y}_{k,c},\mathbf{y}_{l,c})
+
\gamma,
0
\right)
\Big],
\end{flalign*}
where $C$ is the number of training quadruplets, and $\alpha$, $\beta$, and $\gamma$ are nonnegative margin parameters. The first two terms encourage the reciprocal Alice--Bob and Bob--Alice embeddings to be closer than the corresponding Alice--Eve and Bob--Eve embeddings. The third term further separates the legitimate reciprocal pair from the pair of eavesdropper observations.

The channel-informed loss penalizes disagreement between the estimated channel taps produced by the channel head and the reference channel taps associated with each received probe. In this work, we use normalized mean-squared error (NMSE) over the four links i.e., Alice--Bob, Bob--Alice, Alice--Eve, Bob--Eve links for each training quadruplet
\begin{flalign*}
\mathcal{L}_{\mathrm{Channel}}
=
\frac{1}{4C}
\sum_{c=1}^{C}
\sum_{r\in\{i,j,k,l\}}
\frac{
\left\|
\widehat{\mathbf{h}}_{r,c}
-
\mathbf{h}_{r,c}
\right\|_2^2
}{
\left\|
\mathbf{h}_{r,c}
\right\|_2^2
+
\epsilon
}
\end{flalign*}
where $\epsilon>0$ is a small constant used for numerical stability. Unlike the metric-learning term, which only constrains relative distances in the embedding space, the channel-informed term explicitly guides the network toward representations that preserve information about the multipath channel.

% After training, only the feature vector $\mathbf{y}_{r,p}$ is used for key generation. The channel-estimation head is used as a training-time regularizer and is not required during deployment. The continuous feature vectors generated by the shared embedding head are quantized into binary vectors in the following stage, where bit disagreement ratio (BDR), reconciliation rate, and privacy-amplified key randomness are used to evaluate the final PKG performance.

\section{Reconciliation and Privacy Amplification}

After the neural network has produced the binary output, Alice and Bob obtain correlated but not necessarily identical binary feature vectors, denoted by $\mathbf{y}_i,\mathbf{y}_j\in\{0,1\}^{W}$. As in \cite{11443725}, residual bit disagreements are corrected using a Reed--Solomon (RS) error-correction-based reconciliation scheme \cite{9598159}. Alice partitions $\mathbf{y}_{i}$ into $K=B/Z$ symbols and encodes them using an $\mathrm{RS}(N,K)$ code, where $N\leq 2^{Z}-1$. She transmits only the $N-K$ parity symbols over the public channel. Bob uses these parity symbols and his own sequence $\mathbf{y}_{j}$ to correct symbol errors and obtain $\widehat{\mathbf{y}}_{j}$. Reconciliation is successful when $\widehat{\mathbf{y}}_{j}=\mathbf{y}_{i}$. An $\mathrm{RS}(N,K)$ decoder can correct up to $\lfloor (N-K)/2\rfloor$ unknown symbol errors.

Because the parity symbols are public, Eve is assumed to observe the reconciliation information and may attempt to reconcile her own quantized observations using the same RS code. Therefore, the code rate is selected to provide enough correction capability for Alice and Bob while limiting the side information available to Eve. After successful reconciliation, Alice and Bob apply privacy amplification by hashing the reconciled bit sequence.% with the SHA-3 hash function
% \begin{flalign*}
%     \mathbf{key}
%     =
%     \mathrm{SHA3}\left(\mathbf{y}_{i}\right)
%     =
%     \mathrm{SHA3}\left(\widehat{\mathbf{y}}_{j}\right),
%     \qquad
%     \mathbf{key}\in\{0,1\}^{512}
% \end{flalign*}
% The resulting $512$-bit digest is used as the final crypto key.
% Remove any mention of SHA (Page 4 – “by hashing the reconciled bit sequence with the SHA-3 hash function key = SHA3 (yi) = SHA3 (byj) , key ∈ {0, 1}512.  The resulting 512-bit digest is used as the final key.”   Which is edited to “hashing the reconciled binary feature vector with privacy amplification to produce a key.”  Interestingly this variable Key is not used after this on page 4 so she read that as not having any technical impact and could be replaced generically.  

\section{Performance Evaluation}

% \begin{figure} [t]
% \centering
% \includegraphics[width=\columnwidth]{Images/PowderDenseDeployment-withEve.png}
% \caption{POWDER outdoor deployment for test and evaluation of PHY key generation.}
% \label{fig:DenseDeployment}
% \end{figure}

% \begin{figure} [t]
% \centering
% \includegraphics[width=\columnwidth]{Images/Eve-B210.png}
% \caption{Eve node setup at scenario 1 eavesdroping probes from USTAR-Moran (Alice-Bob) dense nodes.}
% \label{fig:OTALab}
% \end{figure}

% \begin{figure} [t]
% \centering
% \includegraphics[width=0.8\columnwidth]{Images/PowderOTALab.pdf}
% \caption{POWDER indoor OTA setup.}
% \label{fig:OTALab}
% \end{figure}

\begin{figure} [t]
\centering
\includegraphics[width=0.7\columnwidth]{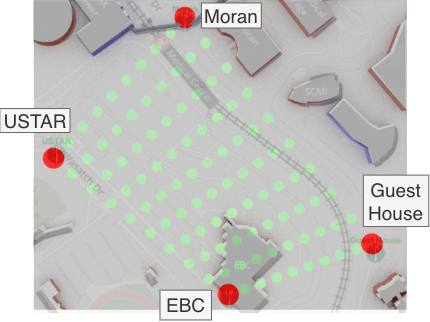}
\caption{POWDER outdoor SDR testbed setup (red dots) and Sionna-RT digital twin setup (green dots) used for training dataset augmentation, test, and evaluation of the proposed PKG protocol.}
\label{fig:DenseDeployment}
\end{figure}

% \begin{figure} [t]
% \centering
% \includegraphics[width=1\columnwidth]{Images/dataset_composition_protocol.png}
% \caption{Dataset composition for training, validation, and testing of our channel informed neural network using a mix of synthetically generated data, indoor, and outdoor OTA data}
% \label{fig:DatasetComposition}
% \end{figure}

\subsection{Experimental Setup}  

% Reviewer comment: The paper describes the framework as suitable for resource-constrained tactical-edge devices. Reporting the trained model size, inference latency, memory requirements, and computational platform would help support this practical claim.

% Reviewer comment: I might have liked to have a little more discussion on deployment scenarios. Given the reported reconciliation rates, one might expect to have to run the protocol more than once. Nothing is reported about the amount of time this might take, which could be a determining factor in some deployed conditions. It would also be nice to have a more comprehensive understanding of how this approach compares with other approaches such as those that pre-compute the channel features. While there is certainly some variability in the environmental test conditions, it's not clear how well-tuned the training needs to be to the testing conditions. Given the strict length constraints, however, I understand such expanded discussions would be difficult to add.

We evaluate the proposed channel-informed learning PKG framework using indoor and outdoor over-the-air (OTA) SDR measurements. 
To augment the amount of training data, we use Sionna-RT for simulating the outdoor OTA environment in POWDER and define a $10 \times 10$ spatial grid, as seen in Fig. \ref{fig:DenseDeployment}, of virtual Alice/Bob/Eve node locations. We randomly draw $500$ Alice--Bob--Eve triplets and compute the corresponding channel taps using the ray tracer. These taps are then used as an input parameter in the GNU Radio channel model block to run a software-in-the-loop simulation of channel probing. We consider $15$ variations of SNR uniformly drawn from $5$ dB to $30$ dB. In this way, we generate a synthetic dataset containing $7500$ spectrogram and channel quadruplets under controlled propagation and noise conditions. 

The indoor and outdoor OTA datasets are collected using PAWR's POWDER platform \cite{10.1145/3411276.3412204}. For the OTA dataset, channel taps are acquired via LS channel estimation. The indoor OTA lab consists of 4x NI USRP B210 and 4x NI USRP X310 SDRs separated by approximately $6.4$ m. We use this indoor testbed to collect BPSK GLFSR channel-probes over $10$ Alice--Bob--Eve topology configurations. In each configuration, Alice and Bob exchange $100$ probes in TDD mode, while Eve passively records the transmitted probes creating $1000$ spectrogram and channel quadruplets for indoor training and testing. In the outdoor OTA evaluation, we use POWDER's radios \cite{POWDER-hardware} as seen in Fig. \ref{fig:DenseDeployment}. Specifically, we use the SDRs located at USTAR, EBC, and Guest House, with coordinates $(40.76852,-111.84045)$, $(40.76702,-111.83807)$, and $(40.76749,-111.83607)$, respectively. We collect channel probing data over $2$ Alice--Bob--Eve topology configurations by rotating the roles of the three radios and compare with results shown in \cite{11443725}. 

For both the simulated and OTA experiments, we set the GLFSR BPSK sequence $N_p = 256$ and number of channel taps $P = 16$. The carrier frequency is set to $f_c=3.5$ GHz and the receiver sampling rate is set to $1$ MHz. %Each channel probe consists of a known BPSK-modulated GLFSR sequence transmitted as a complex baseband waveform. 
For each received channel probe, the IQ samples $N_p+P=256+16=272$ are RMS-normalized and processed into spectrograms using a window of length $N_F=32$, hop size $R=N_F\times0.5=16$, for a total of $M=1+((256+16-32)/16)=16$ segments resulting into spectrogram inputs of size $32\times16\times2$. The resulting spectrograms are organized into quadruplets $\left<\mathbf{S}_{i},\mathbf{S}_{j},\mathbf{S}_{k},\mathbf{S}_{l}\right>$, corresponding to Alice--Bob, Bob--Alice, Alice--Eve, and Bob--Eve observations, respectively. 
%Each quadruplet is paired with channel-tap targets obtained either from least-squares channel estimation in the measured OTA datasets or from the Sionna-RT-generated channel taps in the synthetic dataset.  
The channel-informed RNN encoder is trained using directly IQ spectrograms and produces two outputs: a $W=128$ binary feature vector that can be used directly as an encryption key and a $P=16$-tap complex channel estimate. The $5.02$ MiB RNN model contains $1,290,656$ parameters which are optimized with the combined channel-informed metric-learning objective. We set $\alpha$, $\beta$ and $\gamma$ to $0.5$, $0.5$, and $0.1$ respectively, the channel weight $\lambda_{ch}$ to $0.3$, and trained the neural network using the RMSprop optimizer and a learning rate of $0.0001$. The average inference time on a NVIDIA-H200 GPU and linux system over 100 runs is 2.56 ms and 0.04 GFLOPs per sample. Indoor OTA and simulation data containing $8500$ quadruplets are split into $70\%$ training, $15\%$ validation, and $15\%$ testing 
%across indoor OTA and synthetic sources to train and validate in both controlled synthetic performance and generalization to measured indoor while making use of 
while the entire outdoor OTA dataset containing 200 quadruplets was reserved for outdoor testing purposes only.

% Reviewer comment: Please clarify how the indoor training, validation, and test sets were partitioned. It is important to indicate whether probes from the same topology, radio configuration, or collection session appear in multiple subsets. A topology-disjoint split would provide stronger evidence that the model is not memorizing fixed propagation or hardware characteristics. (We already have the topology description as three nodes Alice, bob, eve where alice bob is bidirectional and one way to eve)

%The final dataset consists of 
%$7{,}500$ quadruplets from the synthetic Sionna-RT/GNU Radio dataset, 
%$1000$ quadruplets from the POWDER indoor OTA dataset, and $187$ quadruplets from the POWDER outdoor OTA dataset. Each quadruplet contains Alice--Bob, Bob--Alice, Alice--Eve, and Bob--Eve observations and is paired with channel-tap targets obtained from 
%either Sionna-RT ray tracing or 
%LS channel estimation.

\subsection{Experimental Results}

% Reviewer comment: Since Eve-related observations are explicitly included as negative examples during training, additional discussion of generalization to previously unseen eavesdropper locations and hardware would be valuable. Testing with Eve at different distances, including locations closer to Alice or Bob, would further strengthen the security evaluation.
% Perhaps add results on simulated scenarios with eavesdropper at different locations (including close to the receiver). Perhaps having a BDR curved line plot that shows closer BDR as eve gets closer.

% FIX IMAGE
% Make it centered/bigger. Add reference for sinusoid (RNN [18], CI-RNN, CI-RNN RT Augmentation)
% Remove node comparison
% CHANGE NAMES IN TEXT, FIGURES, AND TABLES
\begin{figure}[t]
\centering
\includegraphics[width=1\columnwidth]{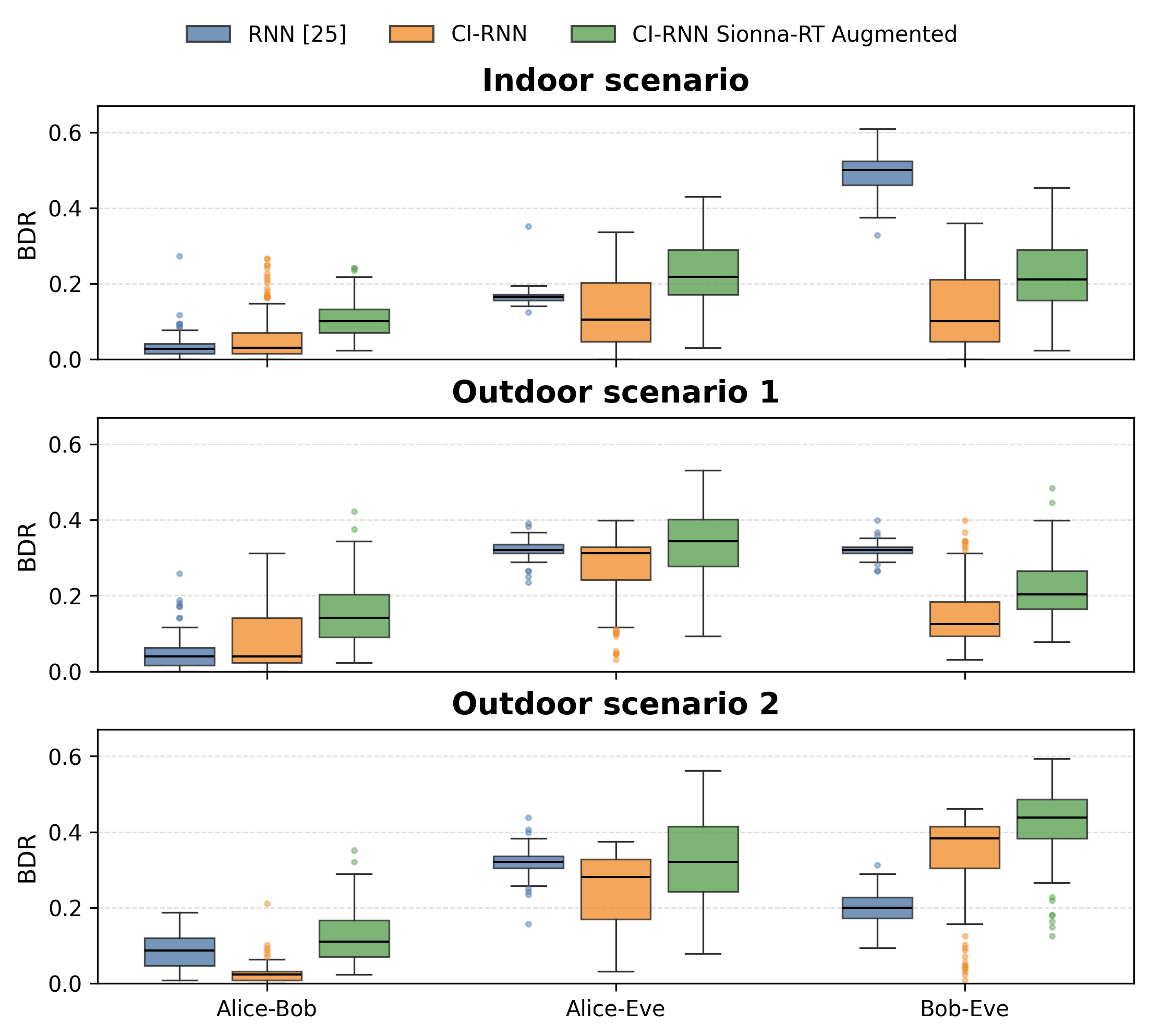}
\caption{$\rm{BDR}$ distributions across indoor and outdoor Alice--Bob--Eve topology configurations for RNN \cite{11443725}, CI-RNN, CI-RNN Sionna RT Augmented models.}
\label{BDR2}
\end{figure}

We evaluated the trained channel-informed feature extractor on one indoor and two outdoor OTA scenarios. The indoor scenario contains $150$ channel probe exchanges, while outdoor scenarios $1$ and $2$ contain $95$ and $92$ probe exchanges, respectively. We also compare our results with our previous baseline work in \cite{11443725} which relies on simple tone-based channel probes and the same backbone RNN architecture on the same exact three (indoor, outdoor) OTA scenarios considering only the results when generating keys of size $W=128$. 
% Reviewer Comment: which relays on simple tone-based -> relies
For each scenario, we compute the bit disagreement ratio (BDR) for any two binary feature vectors as
\begin{flalign*}
{\rm{BDR}}_{i,j,c} = \frac{1}{W}\sum_{w=1}^{W} y_{i,c,w} \oplus y_{j,c,w}
% \label{bdreq}
\end{flalign*}
between the generated Alice--Bob, Alice--Eve, and Bob--Eve observations.

Fig.~\ref{BDR2} shows the BDR distributions across the three OTA test scenarios and three PKG learning models i.e., 1) RNN in \cite{11443725}; 2) channel-informed (CI) RNN trained on indoor OTA data; 3) CI-RNN trained on both indoor OTA data and Sionna-RT augmented data. These results show that across all scenarios, Alice-Bob BDR consistently produces more similar binary vectors for reciprocal legitimate channels than for eavesdropper channels.

Across the generated keys we evaluated key diversity using the unique key rate as well as hamming distance across the generated binary feature vectors for each model and OTA scenario by counting the number of unique keys divided by the total number of generated keys. Even though the RNN model in \cite{11443725} offered the lowest Alice-Bob BDR results, it also reported the lowest unique key rates of $0.21$, $0.29$, and $0.33$. The CI-RNN  trained only on indoor OTA data improves these rates to $0.37$, $0.63$, and $0.57$ for the indoor, outdoor 1, and outdoor 2 scenarios, respectively. Training the CI-RNN on both indoor-OTA and Sionna-RT augmented data further improves the unique-key rates to $0.94$, $0.99$, and $0.99$ across the same three scenarios indicating that RT-based data augmentation helps the proposed channel-informed learning PKG framework  generate more unique keys.

We evaluate reconciliation using a Reed--Solomon code $\mathrm{RS}(24,16)$ and $\mathrm{RS}(32,16)$ over $Z=8$-bit symbols. Since the binary feature vector has length $W=128$, each candidate key is represented by $K=B/Z=16$ symbols. We try two encoders and expand to $N=24$ $N=32$ symbols, corresponding to code rates of $R=K/N=0.66$ and $R=0.5$, respectively. Reconciliation is successful when Bob's decoded binary sequence matches Alice's binary vector.

% EXTEND TABLE FOR RECONCILIATION RESULTS FOR OTHR NETWORKS + Different Rates
Table~\ref{table:reconciliation_rates} reports the reconciliation rates for Alice--Bob, Alice--Eve, and Bob--Eve pairs. 
%In the indoor OTA scenario, Alice--Bob achieves a reconciliation rate of $0.36$, while Alice--Eve and Bob--Eve achieve only $0.0667$ and $0.0933$, respectively. In outdoor scenario 1, Alice--Bob reconciles at $0.2421$, while Alice--Eve does not reconcile and Bob--Eve achieves $0.0421$. In outdoor scenario 2, Alice--Bob achieves $0.3587$, while both Eve-related pairs have zero reconciliation success. 
% RECONCILIATION DISCUSSION
As expected, lower-rate RS codes result in higher reconciliation rates for Alice-Bob at the cost of Eve reconciling as well. Reconciliation rates for the proposed CI-RNN method are lower than the rates reported in \cite{11443725}, which is expected due to lower uniqueness among the generated keys. 
% we get higher reconciliation rates for Alice-Bob at  provides higher reconciliation success for reciprocal legitimate observations than for Eve-related observations, although the legitimate reconciliation rate remains limited by the residual Alice--Bob BDR for CI-RNN due to incremental key generation randomness.

\begin{table*}[t]
\centering
\caption{Reconciliation rates (RR) for $Z=8$ bits per symbol and code rates $R=0.66,0.50$ for $\mathrm{RS}(24,16)$ \& $\mathrm{RS}(32,16)$ respectively.}
\label{table:reconciliation_rates}
\begin{tabular}{|l|c|ccc|ccc|}
\hline
\multicolumn{2}{|c|}{\textbf{Code Rate}}  &  \multicolumn{3}{|c|}{$R = 0.66$ 
%| $\mathrm{RS}(24,16)$
}  &  \multicolumn{3}{|c|}{$R = 0.50$ 
%| $\mathrm{RS}(32,16)$
}  \\
\hline
\textbf{Scenario} & \textbf{Model} & RR$_{\mathrm{AB-BA}}$ & RR$_{\mathrm{AB-AE}}$ & RR$_{\mathrm{BA-BE}}$ & RR$_{\mathrm{AB-BA}}$ & RR$_{\mathrm{AB-AE}}$ & RR$_{\mathrm{BA-BE}}$ \\
\hline
                & RNN \cite{11443725} & $0.67$ & $0.00$ & $0.00$ & $0.93$ & $0.00$ & $0.00$ \\
Indoor      & CI-RNN & $0.61$ & $0.24$ & $0.18$ & $0.80$ & $0.48$ & $0.48$ \\
                & CI-RNN (Sionna-RT Augmented) & $0.08$ & $0.00$ & $0.02$ & $0.36$ & $0.06$ & $0.09$ \\
                \hline
                & RNN \cite{11443725} & $0.5$ & $0.00$ & $0.00$ & $0.87$ & $0.00$ & $0.00$ \\
Outdoor 1   & CI-RNN & $0.47$ & $0.03$ & $0.01$ & $0.71$ & $0.06$ & $0.21$ \\
                & CI-RNN (Sionna-RT Augmented) & $0.03$ & $0.00$ & $0.00$ & $0.24$ & $0.00$ & $0.04$ \\
                \hline
                & RNN \cite{11443725} & $0.18$ & $0.00$ & $0.00$ & $0.43$ & $0.00$ & $0.06$ \\
Outdoor 2   & CI-RNN & $0.85$ & $0.03$ & $0.05$ & $0.96$ & $0.11$ & $0.11$ \\
                & CI-RNN (Sionna-RT Augmented) & $0.08$ & $0.00$ & $0.00$ & $0.35$ & $0.00$ & $0.00$ \\
\hline
\end{tabular}
\end{table*}

% Reviewer comment: The NIST randomness-test methodology should be described in greater detail. Please clarify whether the 128-bit keys were tested individually or concatenated, the total number of tested bits, the significance threshold, the test parameters, and the number of successfully reconciled keys used in each scenario. 
% "prior to privacy amplification" stated below

The following nine tests from the NIST Statistical Test Suite \cite{NIST} were performed on successfully reconciled keys prior to privacy amplification: Approximate Entropy, Cumulative Sums, Frequency Within Block, Longest Run of Ones in a Block, Monobit, Non-Overlapping Template Matching, Random Excursion, Runs, and Serial with respective passing scores.
% Reviewer comment: With  - Non-Overlapping Template Matching, Random Excursion, runs and Serial, should runs be capitalized?
% The corresponding pass scores are reported in Table~\ref{table:nist_results}. 
% \begin{table}[t]
% \centering
% \caption{NIST randomness test pass scores for successfully reconciled keys before privacy amplification.}
% \label{table:nist_results}
% \begin{tabular}{lc}
% \hline
% NIST test & Pass score \\
% \hline
% Approximate Entropy & $0.83$ \\
% Cumulative Sums & $1.00$ \\
% Frequency Within Block & $0.70$ \\
% Longest Run of Ones in a Block & $0.05$ \\
% Monobit & $0.47$ \\
% Non-Overlapping Template Matching & $0.54$ \\
% Random Excursion & $0.97$ \\
% Runs & $0.61$ \\
% Serial & $0.70$ \\
% \hline
% \end{tabular}
% \end{table}
The results indicate that 100\% of the successfully reconciled keys meet the evaluated NIST randomness for CI-RNN generated keys while generated keys in \cite{11443725} fail the Random Excursion Variant test.

% PRIVACY AMPLIFICATION AND RANDOMNESS RESULTS
After successful reconciliation, Alice and Bob apply privacy amplification by hashing the reconciled binary feature vector assumed to be public. %with SHA-3 

\section{Conclusions and Future Work}

This paper introduced a channel-informed neural framework for physical-layer key generation directly from raw IQ measurements. The proposed approach extends prior AI-assisted PKG by combining structured channel probing, multipath-channel supervision, and ray-traced data augmentation within a unified learning pipeline. Rather than optimizing key-feature similarity alone, the proposed multi-task architecture jointly learns reciprocity-preserving binary features and an auxiliary representation of the wireless channel. This design grounds the learned key material in measurable propagation structure while retaining the lightweight and decentralized operation required for edge key establishment.

The framework was evaluated using indoor and outdoor over-the-air SDR measurements collected on the POWDER testbed, together with Sionna-RT-generated propagation scenarios used for training augmentation. Across all evaluated scenarios, the proposed channel-informed RNN produced lower bit disagreement for reciprocal Alice–Bob observations than for Eve-related observations. The most significant improvement was observed in key diversity. % while the prior sinusoidal-probe RNN baseline achieved unique-key rates of 0.21, 0.29, and 0.33 across the indoor and two outdoor scenarios, the proposed channel-informed model trained with both measured and ray-traced data increased these rates to 0.94, 0.99, and 0.99, respectively. 
Successfully reconciled channel-informed keys also passed the selected NIST randomness tests prior to privacy amplification. %Following reconciliation, SHA-3 hashing produces 512-bit cryptographic keys for secure data exchange.

The results expose an important agreement–diversity tradeoff. Increasing the sensitivity of the learned representation to channel-dependent variations substantially improves key diversity, but it also increases residual Alice–Bob disagreement and reduces reconciliation success under the evaluated Reed–Solomon configurations. This tradeoff identifies a central design challenge for learning-assisted PKG: improving the entropy and uniqueness of generated keys without sacrificing reliable agreement between trusted users. Future work will investigate reconciliation-aware training objectives, adaptive quantization and error-correction strategies, alternative channel-informed losses, and broader validation under mobility, interference, larger network topologies, and active adversarial attacks.

\balance
\bibliographystyle{IEEEtran}
\bibliography{bib2}

% that's all folks
\end{document}